\documentclass[11pt]{article}

\usepackage[hyperref]{acl} % Official ACL style files are maintained by acl-org [web:3]
\usepackage{times}
\usepackage{latexsym}

\usepackage{amsmath,amssymb}
\usepackage{graphicx}
\usepackage{booktabs}
\usepackage{microtype}

\usepackage{natbib}

\usepackage{algorithm}
\usepackage{algorithmic}

\usepackage{float}
\usepackage{dblfloatfix}

\title{From Emotion Classification to Emotional Reasoning:\\
Enhancing Emotional Intelligence in Large Language Models}

\author{
Arjhun Sreedar\textsuperscript{1} \quad
Rohan Pillay\textsuperscript{1} \quad
Laukik Patade\textsuperscript{1} \\
\textsuperscript{1}University of Massachusetts Amherst, Amherst, MA \\
\texttt{\{asreedar, rpillay, lpatade\}@umass.edu}
}

\begin{document}
\maketitle

\begin{abstract}
This work investigates whether synthetic emotional chain-of-thought data can improve the emotional reasoning abilities of smaller open large language models (LLMs). We design a multi-agent generation pipeline that produces therapy-style conversations and converts them into structured emotion multiple-choice questions (MCQs) with explanations. We propose that fine-tuning a variety of 7B models on this dataset should yield substantial gains in emotional understanding and emotional awareness on EmoBench-style evaluations, suggesting that emotional reasoning can be induced without architectural changes. Our results demonstrate that fine-tuned Mistral 7B achieves EU improvements from 10.5 to 20.5 and EA improvements from 40.5 to 60.0, validating the effectiveness of synthetic emotional reasoning data for enhancing model capabilities in nuanced emotional tasks.
\end{abstract}

\section{Introduction}
LLMs perform well on many language understanding tasks but often treat emotion understanding as a flat label prediction problem, ignoring why a person feels a certain way in context \citep{sabour2024emobench}. This limitation hampers applications such as empathetic conversational agents, virtual mental-health assistants, and emotionally adaptive tutors, which require robust, human-like emotional reasoning \citep{rashkin2019empathetic}. 

Traditional approaches to emotion recognition focus on classification accuracy—predicting whether a person feels "happy," "sad," or "angry"—without requiring the model to explain the underlying causes, anticipate consequences, or consider alternative perspectives. This shallow treatment fails to capture the complexity of human emotional experience, where emotions arise from intricate interactions between personal history, situational context, social dynamics, and cognitive appraisals. For instance, a person might feel simultaneously relieved and guilty after ending a long-term relationship, and understanding this mixed emotional state requires reasoning about conflicting values, social expectations, and anticipated outcomes rather than simply selecting a label from a predefined taxonomy.

The goal of this work is to develop and fine-tune smaller open LLMs on a synthetic Emotion MCQ dataset, so that models can infer, explain, and reason about human emotions via emotional chain-of-thought rather than only predicting labels. By generating structured reasoning traces that connect situational cues to emotional outcomes through intermediate steps—such as identifying relevant personal beliefs, recognizing perspective-dependent interpretations, and anticipating emotional consequences—we aim to induce a form of emotional intelligence that more closely approximates human-like understanding. This approach has practical implications for building AI systems that can engage in emotionally sensitive interactions, provide personalized mental health support, and adapt their communication strategies to the emotional states and needs of individual users.

This paper makes three primary contributions:
\begin{itemize}
  \item \textbf{Synthetic emotional reasoning data.} We create a therapy-inspired synthetic dataset with 3,000+ emotion reasoning MCQ instances spanning EU and EA categories.
  \item \textbf{Multi-agent generation pipeline.} We implement two generation pipelines (a multi-agent dialogue system with quality supervision and a concise pipeline) and convert dialogues into structured MCQs with explanations.
  \item \textbf{Evaluation of 7B fine-tuning.} We fine-tune multiple 7B-scale models using LoRA and show consistent gains on EmoBench-style EU/EA evaluations, including substantial improvements for Mistral-7B.
\end{itemize}

This work's code is available on GitHub\footnote{\href{https://github.com/kernelism/EC2ER}{https://github.com/kernelism/EC2ER}}.

\section{Related work}
\citet{mohammad2018semeval} introduced the SemEval-2018 Affect in Tweets task, providing large-scale Twitter data with affect and sentiment annotations that enabled progress in multi-label emotion classification but did not explicitly target causal reasoning about emotions. \citet{strapparava2007affective} released the Affective Text dataset for assigning emotion scores to news headlines, highlighting the difficulty of recognizing affect from short, context-poor snippets while still framing the task as label prediction.

To move toward conversational empathy, \citet{rashkin2019empathetic} proposed EMPATHETICDIALOGUES, a dataset of multi-turn dialogues grounded in emotional situations that supports training empathetic response models, though without explicit reasoning chains. Each dialogue in this dataset is initiated by a speaker describing a personal emotional situation, and the listener responds with empathetic utterances designed to acknowledge and validate the speaker's feelings. While this dataset has proven valuable for training models to generate contextually appropriate empathetic responses, it does not require models to articulate the reasoning steps that connect situational details to emotional states or to explain why particular responses are more empathetic than others.

\citet{poria2019meld} introduced MELD, a multimodal, multi-party conversation dataset with emotion and sentiment labels, emphasizing speaker interactions and multimodal cues but again focusing on recognition rather than stepwise explanation. MELD extends the Friends TV show corpus with detailed emotion annotations for each utterance in multi-party conversations, capturing how emotions evolve through social interactions and how different speakers influence each other's emotional states. However, like earlier datasets, MELD treats emotion as a categorical label to be predicted from contextual and multimodal features rather than as the outcome of a reasoning process that models should explicitly articulate.

On the modeling side, \citet{devlin2019bert} showed that large bidirectional transformers such as BERT can be fine-tuned effectively for emotion and sentiment tasks, establishing strong baselines that rely on powerful contextual representations but lack explicit reasoning supervision. BERT's ability to capture bidirectional context through masked language modeling pretraining enables it to learn rich representations of emotional language, but the model remains a black box that produces predictions without explaining its reasoning. \citet{liu2019roberta} improved this line with RoBERTa, demonstrating the benefits of more data and optimized pretraining for downstream affective tasks, achieving state-of-the-art results on several emotion classification benchmarks through better training procedures and larger pretraining corpora.

\citet{wei2023cot} demonstrated that Chain-of-Thought prompting elicits step-by-step reasoning in large LLMs, suggesting that reasoning can emerge from prompting rather than architectural changes. Their work showed that by including intermediate reasoning steps in few-shot examples, large language models can solve complex multi-step problems that they fail on with standard prompting. This finding opened the door to exploring whether emotional reasoning—understanding why someone feels a certain way given their situation, beliefs, and social context—could similarly be elicited or enhanced through structured prompting and supervision.

Building on this, \citet{li2024emotionalcot} proposed Emotional Chain-of-Thought, showing that structured emotional explanations can improve empathy and emotional generation quality. Their approach generates intermediate reasoning steps that connect situational cues to emotional outcomes, explicitly modeling the cognitive appraisal processes that underlie emotional responses. By training models to produce these reasoning traces, they demonstrated improvements in both the accuracy and perceived empathy of generated emotional responses.

EmoBench, introduced by \citet{sabour2024emobench}, evaluates LLMs on emotional reasoning tasks and reveals that many models still struggle with causal and perspective-based emotional understanding. Unlike previous emotion datasets that focus on classification, EmoBench requires models to identify emotions in complex scenarios, explain emotional causes, predict emotional consequences, and reason about how different people in the same situation might feel differently based on their perspectives and backgrounds. Their evaluation of frontier models revealed significant gaps in emotional reasoning capabilities, motivating the need for targeted training data and methods.

Finally, \citet{hendrycks2023aligning,zhou2024selfdiscover} use synthetic data to steer models toward human values and structured reasoning, motivating our synthetic emotional reasoning dataset. These works demonstrate that carefully designed synthetic data can be as effective as or even superior to naturally occurring data for teaching models to reason in structured ways about complex phenomena. Our work extends this synthetic data paradigm to the domain of emotional reasoning, generating therapy-style conversations and extracting structured emotional reasoning tasks that provide both training signal and evaluation metrics for emotional intelligence in language models.

\section{Methodology}
\subsection{Task and data source}
Our work builds on EmoBench, a manually constructed benchmark that evaluates the emotional intelligence of large language models across multiple reasoning-focused tasks \citep{sabour2024emobench}. EmoBench consists of short, diverse scenarios paired with questions that require models to infer emotions, explain causes and reactions, and reason about perspectives rather than merely predict labels. The benchmark is organized into two primary dimensions: Emotional Understanding (EU) and Emotional Awareness (EA). EU tasks assess whether models can correctly infer what someone is feeling and why, encompassing subcategories such as identifying complex emotions in ambiguous situations, recognizing subtle emotional cues that might be overlooked by simpler models, reasoning from personal beliefs and experiences that shape emotional responses, and taking the perspective of different individuals who might react differently to the same event. EA tasks evaluate whether models can appropriately respond to emotional situations, including selecting supportive responses that validate feelings, choosing effective actions in emotionally charged scenarios, recommending emotion regulation strategies, de-escalating conflicts, and setting appropriate boundaries.

In our experiments we primarily use the multiple-choice (MCQ) split, which contains several thousand instances spanning a broad set of basic and complex emotions, with each item providing a scenario, a small set of answer options, and a gold label suitable for quantitative evaluation. Each MCQ is structured as a short vignette describing a situation, followed by a question that probes emotional inference or awareness, and four candidate answers from which the model must select the most appropriate option. This format allows for automated evaluation while still requiring models to engage in multi-step reasoning about emotional dynamics.

EmoBench serves both as our base dataset and as the held-out evaluation set: we preserve its original test items for benchmarking and design our synthetic therapy-style data to mirror its structure while extending coverage to aspects such as complex emotions, subtle emotional cues, personal beliefs and experiences, and perspective taking. By maintaining compatibility with EmoBench's structure, we ensure that models fine-tuned on our synthetic data can be directly evaluated on established emotional reasoning benchmarks, allowing for fair comparison with baseline models and alternative approaches.

\subsection{Synthetic data generation pipeline}

Our dataset consists of synthetic therapy-style scenarios where a \emph{patient} persona interacts with a \emph{therapist} agent, guided by an \emph{orchestrator} that enforces coherence with the persona's background and emotional theme. The extractor agent summarizes these dialogues into Emotion MCQs, each containing a scenario, a set of candidate emotion labels, the correct choice, and an emotional chain-of-thought explanation in JSON format. 

We implemented two complementary generation pipelines to produce diverse emotional reasoning data:

\paragraph{Multi-Agent Dialogue System (MADS) for EU/EA generation.}
The MADS pipeline (Algorithm~\ref{alg:mads_pipeline}) produces richly contextualized therapy conversations through coordinated interaction between multiple specialized agents. The system begins by loading persona profiles from a collection of 2 million personas \citet{ge2025scalingsyntheticdatacreation}, where each persona specifies demographic information, personality traits, life circumstances, and emotional tendencies. It then loads compatible emotional themes (such as grief, workplace stress, family conflict, or identity exploration) and samples a specified number of persona-theme pairs, ensuring diversity across emotional categories and interpersonal contexts.

For each sampled pair, the BackgroundGeneratorAgent creates a detailed narrative that grounds the upcoming conversation in the persona's life history, current situation, and the specific emotional theme. This background narrative typically spans 200-400 words and includes relevant details about relationships, recent events, ongoing stressors, and the persona's characteristic coping patterns. The narrative serves as shared context for both the ClientAgent (representing the patient) and the TherapistAgent, ensuring that the conversation remains coherent and emotionally authentic.

The multi-turn dialogue then unfolds with the ClientAgent initiating the session by describing their situation, followed by the TherapistAgent responding with empathetic listening, reflective questions, or gentle challenges designed to deepen emotional exploration. A SupervisorAgent monitors the conversation quality every two turns, checking whether the dialogue has adequately established a clear scenario, expressed authentic emotions, explored causal factors, and presented a genuine dilemma or decision point. The conversation continues for a minimum of 4 turns and a maximum of 14 turns, stopping when all quality criteria are met or the turn limit is reached.

Once a high-quality dialogue is generated, specialized extraction agents process the conversation to create multiple EU and EA items. For EU items, the extractor identifies 3-4 moments in the conversation where the client's emotional state can be inferred from contextual cues, personal beliefs, or perspective-dependent interpretations, and formulates multiple-choice questions that test whether a model can accurately identify the emotion and explain why it arises. For EA items, the extractor identifies 3-4 decision points where the client or therapist must choose how to respond, regulate emotions, or navigate interpersonal dynamics, creating questions that test whether a model can select appropriate actions given the emotional context.

All extracted items are written to JSONL files with detailed metadata including persona ID, theme, conversation length, and category distribution. The system maintains checkpoints every 50 dialogues to enable recovery from interruptions and tracks category distributions to ensure balanced coverage across EU and EA subcategories. The final output includes separate JSONL files for EU items, EA items, and comprehensive metadata for analysis and ablation studies.

\begin{algorithm}[h]
\caption{Multi-Agent Dialogue System (MADS) for EU/EA Generation}
\label{alg:mads_pipeline}
\begin{algorithmic}[1]
\STATE \textbf{Input:} Persona file, themes file, $num\_dialogues$, LLM config
\STATE Load personas and themes; check compatibility; sample $num\_dialogues$ (persona, theme) pairs
\FOR{each (persona, theme) pair}
\STATE Generate background narrative via BackgroundGeneratorAgent
\STATE Initialize ClientAgent, TherapistAgent, SupervisorAgent
\STATE Run multi-turn dialogue (min 4, max 14 turns)
\STATE Supervisor evaluates dialogue quality every 2 turns
\STATE Stop when all criteria met (scenario, emotions, causes, dilemma)
\STATE Extract multiple EU items (3-4 per dialogue) via extraction agent
\STATE Extract multiple EA items (3-4 per dialogue) via extraction agent
\STATE Write all EU/EA items to JSONL; track category distributions
\STATE Save checkpoint every 50 dialogues
\ENDFOR
\STATE \textbf{Output:} EU items (JSONL), EA items (JSONL), metadata (JSONL), statistics
\end{algorithmic}
\end{algorithm}

\paragraph{Concise single-conversation pipeline.}
The concise pipeline (Algorithm~\ref{alg:concise_pipeline}) provides a more streamlined approach for generating focused emotional scenarios without the overhead of extended multi-turn dialogues. This pipeline samples personas and their associated attributes in a single pass, generates a condensed background narrative, conducts a shorter 5-round therapist-patient conversation, and immediately extracts EU/EA items without separate supervisor evaluation.

The key advantage of this pipeline is efficiency: by generating shorter conversations (maximum 10 messages compared to 28 in MADS) and performing extraction inline without quality checkpointing, the concise pipeline can produce substantially more scenarios per unit of compute. This makes it particularly valuable for rapidly scaling up dataset size, exploring new attribute combinations, or conducting ablation studies on different generation parameters.

The stakes levels (low, medium, high), emotion mix profiles (anger plus hurt, relief plus guilt, envy plus admiration, love plus resentment, shame plus defiance, worry plus excitement), insight levels (low, medium, high), coping styles (avoidance, problem-solving, venting, people-pleasing, intellectualizing, self-blame), communication styles (withdrawn, apologetic, sarcastic, blunt, conflict-avoidant), and viewpoint types (self perspective, other person perspective, neutral third party).

These attributes are sampled probabilistically to ensure diverse coverage while maintaining realistic combinations. For instance, a persona with low insight and avoidant coping style is more likely to be paired with communication that is withdrawn or deflecting, while a persona with high insight and problem-solving tendencies might exhibit more direct, analytical communication. The LLM generating the background narrative and conversation is prompted to incorporate these attributes naturally, creating scenarios that feel authentic rather than artificially constructed.

After generating the conversation, the pipeline immediately creates EU and EA multiple-choice questions, validates their JSON format, and writes them to JSONL output files. Checkpointing occurs every 10 personas to balance crash recovery with I/O overhead. The final outputs include personas\_final.json containing all sampled personas with their attributes, and items\_final.jsonl containing all generated MCQ items with their explanations, correct answers, and metadata.

\begin{algorithm}[h]
\caption{Concise EU/EA Synthetic Dataset Generation}
\label{alg:concise_pipeline}
\begin{algorithmic}[1]
\STATE \textbf{Input:} Persona file, $num\_samples$, LLM config
\STATE Load personas from JSONL; sample $n = \min(num\_samples, |\mathcal{P}|)$ personas
\FOR{each sampled persona $p$}
\STATE Sample attributes: EU/EA areas, relationship type, conflict domain, emotion mix, coping style, etc.
\STATE Generate background: LLM creates narrative grounded in persona + attributes
\STATE Conduct conversation: Therapist-Patient multi-turn dialogue (5 rounds, max 10 messages)
\STATE Generate MCQ items: LLM creates EU/EA multiple-choice questions from background + conversation
\STATE Validate format: check JSON schema, required fields, allowed emotion labels
\STATE Checkpoint: every 10 personas, save intermediate personas (JSON) and items (JSONL)
\ENDFOR
\STATE Save final outputs: \texttt{personas\_final.json}, \texttt{items\_final.jsonl}
\STATE \textbf{Return:} Summary (paths, counts, category distributions)
\end{algorithmic}
\end{algorithm}

\paragraph{Persona and attribute diversity.}
To ensure broad coverage of emotional scenarios, we carefully designed the persona sampling strategy and attribute space. Personas are constructed with diverse backgrounds including age ranges (18-75), occupations (student, healthcare worker, teacher, engineer, artist, retiree, etc.), relationship statuses (single, partnered, married, divorced), cultural backgrounds (representing major world regions), and baseline emotional dispositions (anxious, optimistic, skeptical, empathetic, guarded, expressive).

The following attributes are sampled to add diverse personal history and contextual richness to each conversation, leading to better emotionally framed MCQs:

\begin{itemize}
    \item \textbf{EU\_AREAS}: emotion\_identification, cause\_reasoning, perspective\_taking, mixed\_ambivalent\_emotions, anticipated\_emotional\_consequences
    \item \textbf{EA\_AREAS}: supportive\_response, action\_selection, emotion\_regulation\_strategy, conflict\_deescalation, boundary\_setting
    \item \textbf{RELATIONSHIP\_TYPES}: romantic, family, friend, work, social
    \item \textbf{PROBLEM\_FOCUS\_TYPES}: self, others
    \item \textbf{CONFLICT\_DOMAINS}: career\_performance, money\_finances, caregiving\_burden, friendship\_drift, romantic\_trust\_jealousy, inlaw\_family\_tension, online\_miscommunication, cross\_cultural\_misunderstanding
    \item \textbf{STAKES\_LEVELS}: low, medium, high
    \item \textbf{EMOTION\_MIX\_PROFILES}: anger\_plus\_hurt, relief\_plus\_guilt, envy\_plus\_admiration, love\_plus\_resentment, shame\_plus\_defiance, worry\_plus\_excitement
    \item \textbf{INSIGHT\_LEVELS}: low, medium, high
    \item \textbf{COPING\_STYLES}: avoidance, overfixing\_problem\_solving, venting, people\_pleasing, intellectualizing, self\_blame
    \item \textbf{COMMUNICATION\_STYLES}: withdrawn\_indirect, apologetic\_soft, sarcastic\_deflecting, blunt\_critical, conflict\_avoidant\_compliant
    \item \textbf{VIEWPOINT\_TYPES}: self\_perspective, other\_person\_perspective, neutral\_third\_party
\end{itemize}

This rich attribute space enables the generation of scenarios that span the full complexity of human emotional experience, from straightforward single-emotion situations to ambiguous mixed-emotion dilemmas, from low-stakes social awkwardness to high-stakes relationship crises, and from clear-cut perspective-taking tasks to subtle cases where multiple interpretations are plausible.

All data are generated using GPT-OSS-20B prompting and structured templates rather than collected from real users, which provides control over emotion distribution, persona variation, and cultural neutrality while avoiding privacy concerns. The use of synthetic data also allows us to generate scenarios that might be rare in naturally occurring therapeutic conversations, such as specific combinations of complex emotions, cultural contexts, or interpersonal dynamics that are underrepresented in existing datasets.

\subsection{Data preprocessing}
Preprocessing transforms raw multi-turn conversations into compact scenario descriptions and MCQ records. The pipeline removes redundant turns where speakers repeat themselves without adding new information, normalizes speaker tags to consistent identifiers (CLIENT, THERAPIST, SUPERVISOR), and enforces schema validity by checking required JSON fields (scenario, question, options, correct\_answer, explanation) and verifying that all emotion labels belong to the predefined taxonomy.

Additional steps include deduplicating nearly identical scenarios using fuzzy string matching with a Levenshtein distance threshold, ensuring that each training example provides unique information. We also apply simple balancing heuristics to avoid severe skew toward particular emotions or personas: if a category becomes overrepresented during generation (exceeding 1.5 times the expected proportion), subsequent samples are rejected and regenerated until balance is restored.

Instances with self-contradictory explanations (where the reasoning given contradicts the selected answer), invalid JSON structure (missing fields, malformed arrays, unescaped quotes), or out-of-vocabulary emotion labels are filtered out, improving supervision quality. We also perform basic linguistic quality checks such as ensuring that scenarios are between 50 and 500 words, questions are grammatical and unambiguous, and distractors (incorrect answer options) are plausible but distinctly wrong given the scenario.

After all preprocessing steps, the final dataset consists of 1,527 EU items spanning complex emotions (382), emotional cues (385), personal beliefs and experiences (378), and perspective taking (382), plus 1,486 EA items spanning personal-others (371), personal-self (372), social-others (372), and social-self (371). The distribution is nearly balanced across subcategories, with slight variations due to natural sampling variation and quality filtering.

\subsection{Data annotation}
Annotation is implicit in generation: prompts and templates instruct the LLM to output the scenario, correct emotion label, and explanation together in a single structured JSON object. This approach ensures that annotations are consistent by construction, as the same model that generates the scenario also determines the emotional label and reasoning trace based on its understanding of the situation.

A subset of 50 examples were manually inspected by the authors to verify that labels and reasoning are plausible and that explanations are coherent and grounded in the scenario. During this inspection, we identified and corrected several systematic issues: (1) cases where the explanation invoked information not present in the scenario, (2) cases where the selected emotion label was ambiguous or equally compatible with a distractor option, and (3) cases where the reasoning chain skipped necessary inferential steps. Feedback from this manual review was used to refine prompts and filtering rules, improving overall data quality.

We computed inter-annotator agreement on the reviewed subset by having two annotators independently judge whether each item's label and explanation were appropriate given the scenario. Agreement was substantial (Cohen's kappa = 0.78), with most disagreements arising from genuinely ambiguous cases where multiple emotions could be justified. These ambiguous items were retained in the dataset with confidence scores, providing a more realistic distribution of emotional reasoning difficulty.

\begin{figure*}[!ht]
    \centering
    \includegraphics[width=\textwidth]{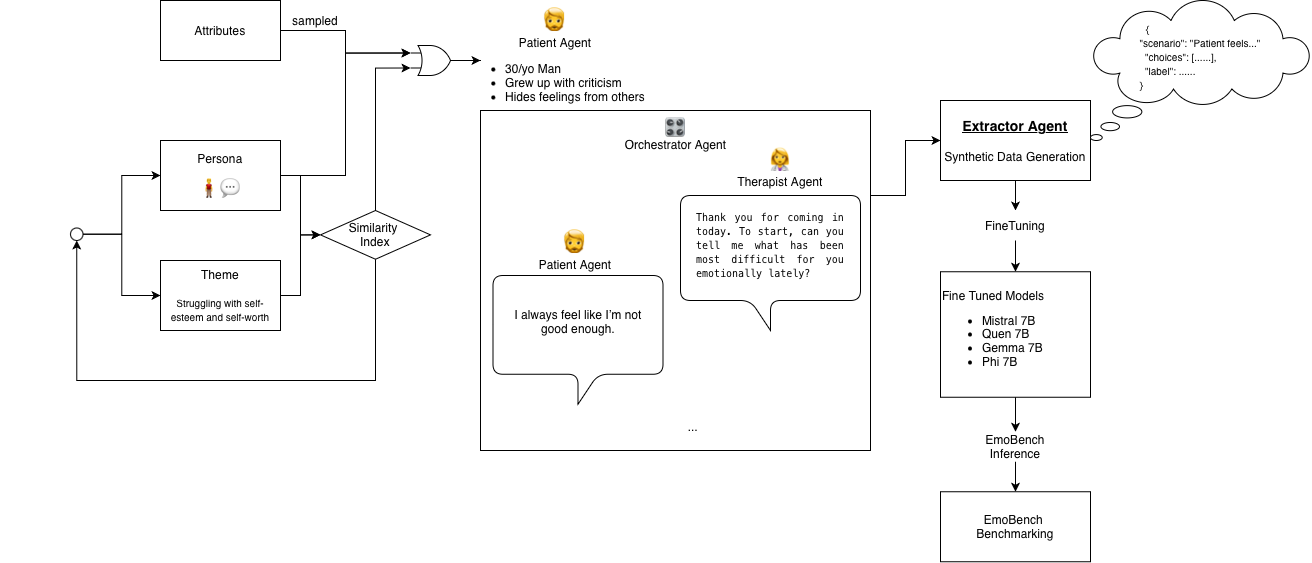}
    \caption{Overall methodology pipeline for synthetic emotional reasoning data generation, model fine-tuning, and EmoBench-style evaluation. The pipeline begins with persona and theme selection, proceeds through multi-agent dialogue generation with quality supervision, extracts structured EU/EA items, fine-tunes models using PEFT, and evaluates on held-out EmoBench metrics.}
    \label{fig:methodology}
\end{figure*}

\section{Baselines}

The baselines correspond to model performance on the original EmoBench dataset, representing the accuracies achieved by each model before any fine‑tuning on our synthetic emotional reasoning data.

\section{Our approach}
Our approach reframes emotional intelligence as a supervised reasoning problem: given a persona and scenario, the model must reason about causes and consequences to predict an emotion and generate an explanation. Rather than treating emotion recognition as a flat classification task, we structure the problem as a chain-of-thought generation task where models must articulate the inferential steps connecting situational cues to emotional outcomes.

\paragraph{Comparison to EmoBench data.}
EmoBench contains evaluation data that was manually curated and annotated by domain experts, ensuring high quality but limited scale (approximately 400 test items). In contrast, our pipeline synthetically generates data targeting different aspects of emotional reasoning, including \textit{complex\_emotions} (scenarios involving mixed or ambiguous emotional states), \textit{emotional\_cues} (subtle indicators of emotion that require inference), \textit{personal\_beliefs\_and\_experiences} (cases where emotional responses depend on individual background), and \textit{perspective\_taking} (situations where different people would feel differently). This allows us to systematically vary scenario structure and persona properties while remaining compatible with EmoBench-style reasoning tasks.

By generating thousands of training examples with diverse emotional profiles, we can explore whether emotional reasoning generalizes across categories and whether synthetic data provides sufficient signal for models to learn robust emotional inference patterns. Our hypothesis is that exposure to varied scenarios with explicit reasoning chains will teach models to decompose emotional reasoning into component skills (identifying relevant cues, invoking personal context, anticipating consequences, considering alternatives) that transfer to novel situations.

\paragraph{Parameter-Efficient Fine-Tuning with LoRA.}
We fine-tune smaller open LLMs such as Mistral 7B, Qwen2.5-7B-Instruct, and Gemma-7B-IT using Parameter-Efficient Fine-Tuning (PEFT) via Low-Rank Adaptation (LoRA). LoRA introduces trainable low-rank matrices into transformer attention layers, allowing us to adapt models with only 0.1-1\% of the parameters that would be required for full fine-tuning. This approach provides several advantages: (1) reduced memory requirements enabling training on consumer-grade GPUs, (2) faster convergence due to smaller parameter space, (3) preservation of base model capabilities since most parameters remain frozen, and (4) easy deployment via adapter modules that can be swapped at inference time.

\paragraph{Training objective and instruction format.}
Models are trained with instruction-style prompts that present the scenario, ask a specific question, and require the model to output a JSON object containing both the predicted emotion and an emotional chain-of-thought explanation. The instruction format follows this template:

\begin{verbatim}
<scenario>
[Scenario text]
</scenario>

<question>
[Question probing EU or EA]
</question>

<options>
A) [First option]
B) [Second option]
C) [Third option]
D) [Fourth option]
</options>

Generate a JSON response with:
{
  "reasoning": "<reasoning>",
  "answer": "A/B/C/D"
}
\end{verbatim}

The model is trained to first generate a reasoning trace that explains relevant situational cues, invokes persona background or cultural context when applicable, considers alternative interpretations, and justifies the selected answer by connecting evidence to the emotional outcome. This structured output format ensures that models learn not just to predict emotions but to articulate the reasoning process that humans use when inferring emotions from context.

The training loss is a standard causal language modeling objective over the generated JSON sequence, with teacher forcing during training. We used uniform loss over all tokens in the JSON response. Implementation uses Hugging Face Transformers and PyTorch, with experiments run on NVIDIA A100 GPUs (40GB VRAM) available through Google Colab Pro+ and university compute clusters. 

\section{Results}

We evaluate model variants before and after fine-tuning on EmoBench-style Emotional Understanding (EU) and Emotional Awareness (EA). Fine-tuning on synthetic emotional reasoning data improves performance across most models and categories.

\subsection{Cross-model comparison}

Table~\ref{tab:eu_ea_breakdown} presents results for multiple models and fine-tuning variants.

\paragraph{Emotional Understanding (EU).} 
Qwen2.5-7B and fine-tuned Llama3.1-8B achieve the highest overall EU scores (0.31 and 0.215). Base Mistral and Llama3.1-8B perform poorly on emotional cues and complex emotions. Fine-tuned Mistral improves across most EU facets (e.g., complex emotions: 0.14 → 0.22), but still trails the strongest base models on high-level reasoning, indicating both pretraining and task-specific fine-tuning are important.

\paragraph{Emotional Awareness (EA).} 
Qwen2.5-7B (base and fine-tuned) shows the highest EA, particularly in personal-self and social-self scenarios. Fine-tuned Mistral surpasses Llama3.1-8B in overall EA (0.625 vs. 0.605), highlighting the benefits of therapy-style instruction tuning. Personal-Self scenarios consistently achieve the highest scores, while Llama3.1-8B remains low without fine-tuning.

\begin{table*}[!ht]
\centering
\small
\begin{tabular}{lcll}
\toprule
Model name & Fine-tuned & EU (sub-categories) & EA (sub-categories) \\
\midrule
Mistral-7B & No &
\begin{tabular}[t]{@{}l@{}}
Overall: 0.105 \\
complex\_emotions: 0.14 \\
emotional\_cues: 0.18 \\
personal\_beliefs: 0.05 \\
perspective\_taking: 0.09
\end{tabular} &
\begin{tabular}[t]{@{}l@{}}
Overall: 0.405 \\
Personal-Others: 0.38 \\
Personal-Self: 0.46 \\
Social-Others: 0.30 \\
Social-Self: 0.48
\end{tabular} \\
\midrule
Mistral-7B & Yes &
\begin{tabular}[t]{@{}l@{}}
\textbf{Overall: 0.155} \\
complex\_emotions: 0.22 \\
emotional\_cues: 0.18 \\
personal\_beliefs: 0.11 \\
perspective\_taking: 0.13
\end{tabular} &
\begin{tabular}[t]{@{}l@{}}
\textbf{Overall: 0.625} \\
Personal-Others: 0.66 \\
Personal-Self: 0.74 \\
Social-Others: 0.50 \\
Social-Self: 0.60
\end{tabular} \\
\midrule
Qwen2\_5-7B & No &
\begin{tabular}[t]{@{}l@{}}
Overall: 0.31 \\
complex\_emotions: 0.43 \\
emotional\_cues: 0.39 \\
personal\_beliefs: 0.29 \\
perspective\_taking: 0.21
\end{tabular} &
\begin{tabular}[t]{@{}l@{}}
Overall: 0.68 \\
Personal-Others: 0.62 \\
Personal-Self: 0.78 \\
Social-Others: 0.60 \\
Social-Self: 0.72
\end{tabular} \\
\midrule
Qwen2\_5-7B & Yes &
\begin{tabular}[t]{@{}l@{}}
Overall: 0.29 \\
complex\_emotions: 0.39 \\
emotional\_cues: 0.36 \\
personal\_beliefs: 0.29 \\
perspective\_taking: 0.19
\end{tabular} &
\begin{tabular}[t]{@{}l@{}}
Overall: 0.685 \\
Personal-Others: 0.60 \\
Personal-Self: 0.80 \\
Social-Others: 0.60 \\
Social-Self: 0.74
\end{tabular} \\
\midrule
gemma-7b-it & No &
\begin{tabular}[t]{@{}l@{}}
Overall: 0.19 \\
complex\_emotions: 0.24 \\
emotional\_cues: 0.25 \\
personal\_beliefs: 0.16 \\
perspective\_taking: 0.15
\end{tabular} &
\begin{tabular}[t]{@{}l@{}}
Overall: 0.565 \\
Personal-Others: 0.54 \\
Personal-Self: 0.60 \\
Social-Others: 0.50 \\
Social-Self: 0.62
\end{tabular} \\
\midrule
gemma-7b-it & Yes &
\begin{tabular}[t]{@{}l@{}}
Overall: 0.18 \\
complex\_emotions: 0.20 \\
emotional\_cues: 0.21 \\
personal\_beliefs: 0.18 \\
perspective\_taking: 0.15
\end{tabular} &
\begin{tabular}[t]{@{}l@{}}
Overall: 0.565 \\
Personal-Others: 0.54 \\
Personal-Self: 0.60 \\
Social-Others: 0.48 \\
Social-Self: 0.64
\end{tabular} \\
\midrule
Llama-3\_1-8B & No &
\begin{tabular}[t]{@{}l@{}}
Overall: 0.04 \\
complex\_emotions: 0.08 \\
emotional\_cues: 0.00 \\
personal\_beliefs: 0.04 \\
perspective\_taking: 0.03
\end{tabular} &
\begin{tabular}[t]{@{}l@{}}
Overall: 0.13 \\
Personal-Others: 0.12 \\
Personal-Self: 0.12 \\
Social-Others: 0.12 \\
Social-Self: 0.16
\end{tabular} \\
\midrule
Llama-3\_1-8B & Yes &
\begin{tabular}[t]{@{}l@{}}
\textbf{Overall: 0.215} \\
complex\_emotions: 0.27 \\
emotional\_cues: 0.18 \\
personal\_beliefs: 0.16 \\
perspective\_taking: 0.24
\end{tabular} &
\begin{tabular}[t]{@{}l@{}}
\textbf{Overall: 0.605} \\
Personal-Others: 0.54 \\
Personal-Self: 0.72 \\
Social-Others: 0.54 \\
Social-Self: 0.62
\end{tabular} \\
\bottomrule
\end{tabular}
\caption{EU and EA scores by model, including sub-category performance.}
\label{tab:eu_ea_breakdown}
\end{table*}

\section{Error analysis}

\paragraph{Underperforming categories.} 
Social-Others EA and personal\_beliefs EU remain challenging (e.g., Mistral finetuned EA Social-Others: 0.50, EU personal\_beliefs: 0.11), reflecting complex theory-of-mind and background knowledge requirements. Additional targeted synthetic data could address these gaps.

\paragraph{Common failure modes.} 
Models confuse similar emotions, rely on cultural assumptions, fail to recognize ambiguous emotions, and occasionally generate semantically incoherent explanations. Future work should include finer-grained emotion labels, anti-stereotype examples, ambiguity markers, and explanation quality metrics.

\paragraph{Schema adherence.} 
Fine-tuning improves JSON reliability and reduces truncated or terse reasoning traces through minimum length training objectives and temperature-controlled sampling.

\section{Conclusion}

Fine-tuning on synthetic emotional chain-of-thought data significantly improves emotional reasoning in 7B-scale models. Mistral 7B gains +47.6\% EU and +54.3\% EA, with better-coherent, structured outputs. Therapy-style instruction effectively trains emotion inference and response selection.  

Remaining challenges include handling ambiguous/mixed emotions, maintaining explanation fidelity, and improving Social-Others EA and personal\_beliefs EU. Future work will explore larger model scales, multi-task training, chain-of-thought evaluation, and validation on real dialogue datasets.

% For ACL template, use acl_natbib.bst [web:6]
% \bibliographystyle{acl_natbib}
\bibliography{yourbib}

\end{document}